\documentclass{article}

\usepackage{arxiv}

\usepackage[utf8]{inputenc} 
\usepackage[T1]{fontenc}    
\usepackage{hyperref}       
\usepackage{url}            
\usepackage{booktabs}       
\usepackage{amsfonts}       
\usepackage{nicefrac}       
\usepackage{microtype}      
\usepackage{lipsum}
\usepackage{graphicx}
\graphicspath{ {./images/} }
\usepackage{makecell}

\title{CascadeV: An Implementation of Würstchen Architecture for Video Generation}

\author{
 Wenfeng Lin \\
 ByteDance \\
 \texttt{linwenfeng.1008@bytedance.com} \\
  \And
 Jiangchuan Wei \\
 ByteDance \\
 \texttt{weijiangchuan@bytedance.com} \\
  \And
  Boyuan Liu \\
 ByteDance \\
 \texttt{liuboyuan@bytedance.com} \\
    \And
 Yichen Zhang \\
 ByteDance \\
 \texttt{zhangyichen.99@bytedance.com} \\
 \And
 Shiyue Yan \\
 ByteDance \\
 \texttt{yanshiyue@bytedance.com} \\'
    \And
 Mingyu Guo \\
 ByteDance \\
 \texttt{guomingyu.313@bytedance.com} \\
}

\begin{document}

\maketitle
\vspace{-1.5cm}
\begin{figure}[htbp]
    \centering
    \includegraphics[width = \textwidth]{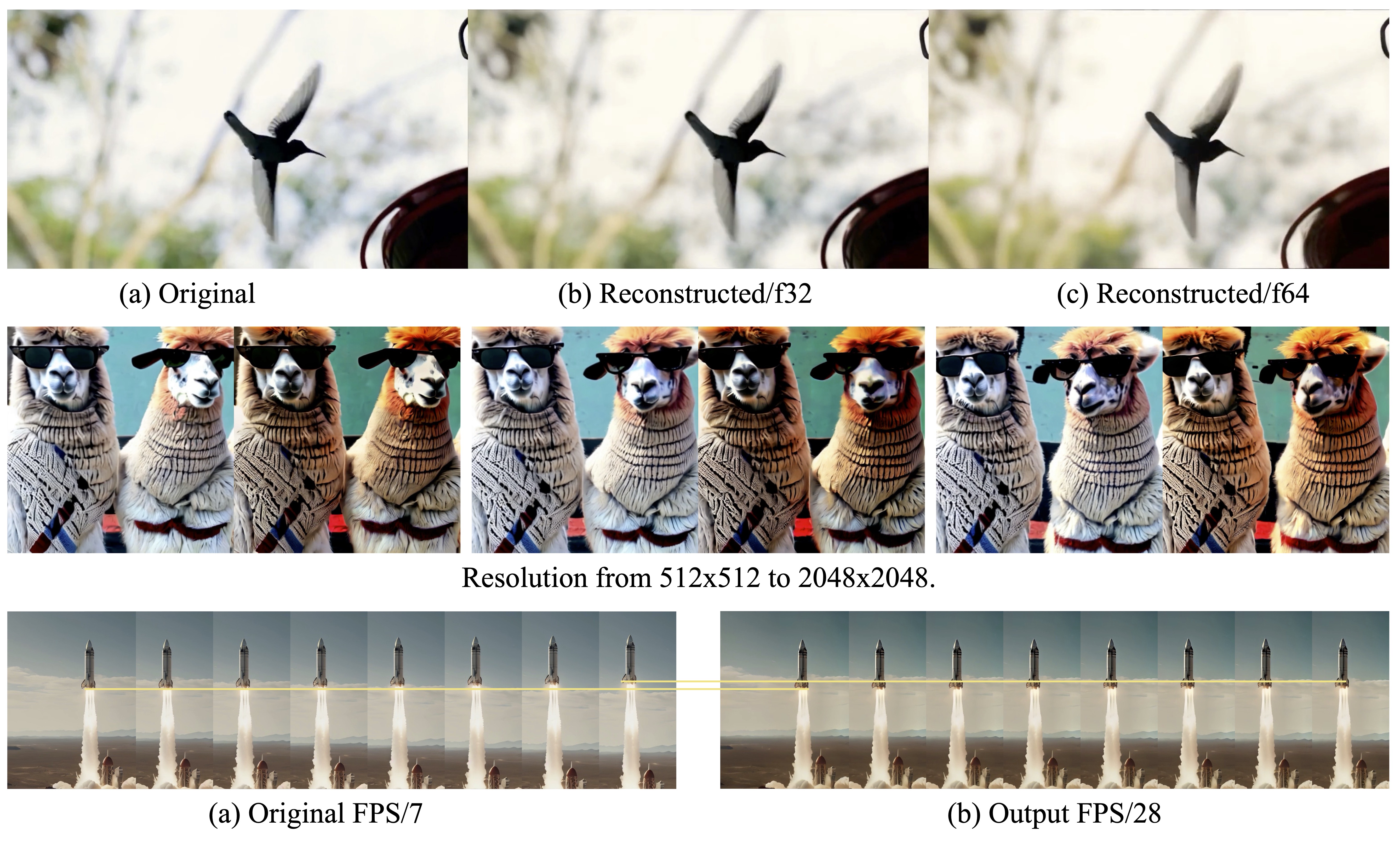}
    \caption{Samples of CascadeV. \textsl{Top}: video reconstruction (of samples from Open-Sora-Plan v1.1.0 \cite{pku_yuan_lab_and_tuzhan_ai_etc_2024_10948109}) with high compression ratio. Even at $64:1$ compression ratios, our model is still able to reconstruct high-frequency details. \textsl{Middle}: 4$\times$ resolution enhancement of Open-Sora-Plan v1.1.0 \cite{pku_yuan_lab_and_tuzhan_ai_etc_2024_10948109} results. \textsl{Bottom}: 4$\times$ FPS improvement of SVD \cite{blattmann2023stable} results. By considering the output of existing T2V models as intermediate results, our model can enhance their resolution and FPS respectively.}
    \label{fig:figure1}
\end{figure}

\begin{abstract}
Recently, with the tremendous success of diffusion models in the field of text-to-image (T2I) generation, increasing attention has been directed toward their potential in text-to-video (T2V) applications. However, the computational demands of diffusion models pose significant challenges, particularly in generating high-resolution videos with high frame rates. In this paper, we propose CascadeV, a cascaded latent diffusion model (LDM), that is capable of producing state-of-the-art 2K resolution videos.
Experiments demonstrate that our cascaded model achieves a higher compression ratio, substantially reducing the computational challenges associated with high-quality video generation. 
We also implement a spatiotemporal alternating grid 3D attention mechanism, which effectively integrates spatial and temporal information, ensuring superior consistency across the generated video frames. 
Furthermore, our model can be cascaded with existing T2V models, theoretically enabling a 4$\times$ increase in resolution or frames per second without any fine-tuning.
Our code is available at \href{https://github.com/bytedance/CascadeV}{https://github.com/bytedance/CascadeV}.
\end{abstract}
\section{Introduction}

\begin{figure}[t]
    \centering
    \includegraphics[width = 0.9\textwidth]{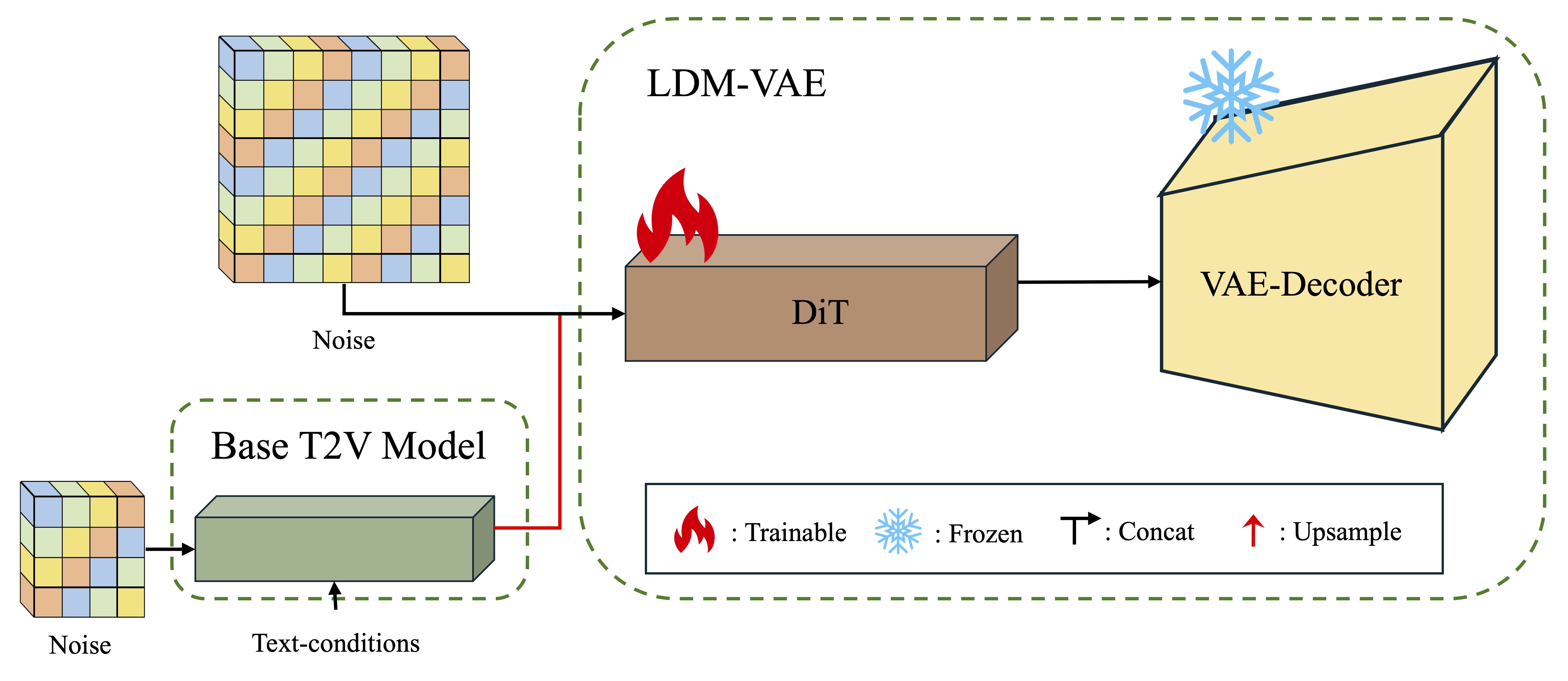}
    \caption{Overall architecture. By cascading the Latent Diffusion Model (LDM), the LDM-VAE can decode the output of the base T2V model with a higher compression ratio.}
    \label{fig:figure2}
\end{figure}

Diffusion models \cite{ho2020denoising, song2020score}, renowned for their powerful generative capabilities, have demonstrated remarkable success across various domains \cite{yang2023diffsound, huang2023make, hoogeboom2023simple, yang2024inf}. 
They have progressively become the paradigm in image generation \cite{esser2024scaling, betker2023improving}, prompting increasing attention to their performance in video synthesis \cite{gupta2023photorealistic, blattmann2023stable}.

However, a significant limitation of diffusion models is their substantial computational demands, often restricting the resolution of image generation to below 1024 pixels. This issue becomes even more pronounced when generating high-quality videos, which inherently possess larger data than static images.
Although leveraging the idea from \cite{rombach2022high}, using Variational Autoencoders (VAE) to compress videos into latent spaces \cite{blattmann2023stable}, allows for model training and inference at lower resolutions, the compression ratio is often quite limited to preserve information integrity \cite{rombach2022high}.

Inspired by the concept of \cite{pernias2023wurstchen}, we propose a cascaded latent space model for T2V generation to address this issue and achieve a spatial compression ratio of $32:1$.
Our approach consists of two primary stages. First, we use a T2V model with a $32:1$ compression ratio as the base model. This stage focuses on semantic information rather than high-frequency details. Subsequently, we employ a VAE based on Latent Diffusion Models (LDM-VAE) to decode the latents generated by the base model, reconstructing high-frequency details to produce 2K resolution videos. 
The LDM-VAE cascades latent Diffusion Transformer (DiT) \cite{peebles2023scalable} with a standard VAE. In the DiT component, we adopt a spatiotemporal alternating grid 3D attention design, which significantly reduces the computational cost of 3D attention while effectively integrating spatial and temporal information. Compared to independently computing spatial and temporal attention, this approach produces more realistic changes between video frames.

Furthermore, we propose that the base model can be trained in the latent space of the standard VAE. This eliminates the need for additional training to obtain a higher compression ratio, decoupling the base model from the DiT model. Consequently, the base model can be replaced by any existing latent video generation models, directly enhancing the quality of videos from current models, as shown in Figure~\ref{fig:figure1}.
Our main contributions are summarized as follows:
\begin{itemize}
    \item We have achieved state-of-the-art(SOTA) performance in generating 2K resolution videos from T2V models.
    \item We propose a cascaded T2V generation model that incorporates 3D attention in the latent space, achieving a $32:1$ compression ratio. This approach significantly reduces the computational resource demands for generating high-quality videos.
    \item Our model can be cascaded with existing T2V models to achieve a 4$\times$ resolution or frames per second (FPS) increase (e.g., Open-Sora-Plan v1.1.0 \cite{pku_yuan_lab_and_tuzhan_ai_etc_2024_10948109}, Stable Video Diffusion (SVD) \cite{blattmann2023stable}).
\end{itemize}

\section{Related Work}

\subsection{Diffusion Model}
Diffusion models \cite{ho2020denoising, song2020score} can generate high-quality results through numerous iterations. To enhance model efficiency, \cite{rombach2022high} significantly reduced computational resource consumption by conducting model training and inference in latent space. \cite{song2020denoising, zhang2022fast, lu2022dpm} proposed training-free sampling algorithms, greatly improving generation speed. By introducing efficient methods for fine-tuning diffusion models, \cite{hu2021lora, zhang2023adding} expand their application prospects. Through cascading approaches, models can produce higher quality results. \cite{pernias2023wurstchen, ho2022cascaded} generate higher resolution images through cascading, while \cite{podell2023sdxl} achieves enhanced detail refinement. Moreover, most existing mature diffusion models utilize U-Net \cite{ronneberger2015u} as their fundamental architecture.  Consequently, \cite{peebles2023scalable} proposes a Transformer-based diffusion model, which offers better scalability and promotes the integration of diffusion models across different tasks.

\subsection{Video Generation}
Generating realistic and vivid videos has long been a challenging task. Numerous approaches based on AR \cite{yan2021videogpt, wu2022nuwa, weissenborn2019scaling}, GAN \cite{yu2022generating, wang2020g3an, tian2020good}, and Transformer \cite{ge2022long, hong2022cogvideo, kondratyuk2023videopoet} architectures have been proposed. With the success of diffusion models, an increasing number of T2V generation works based on diffusion models have emerged in both pixel space \cite{ho2022imagen, hong2022cogvideo, singer2022make} and latent space \cite{blattmann2023align, ge2023preserve, he2022latent, yu2023video}.
By introducing temporal processing modules, several studies \cite{blattmann2023stable,ge2023preserve, singer2022make, pku_yuan_lab_and_tuzhan_ai_etc_2024_10948109} extend pre-trained T2I models to T2V generation. \cite{gupta2023photorealistic, ho2022video}  jointly train on image and video data, enabling models to better capture textual information and significantly improving video quality.
\cite{blattmann2023stable, zhang2023i2vgen} decouple T2V into separate T2I and I2V , leveraging mature T2I models to generate initial frame before rendering the full video. 
\cite{gupta2023photorealistic, blattmann2023align, ho2022imagen} demonstrate cascaded models can generate videos with higher resolution. Similar to these cascaded approaches, our model also adopts a cascaded structure, constructing a VAE with a higher compression ratio through cascading.
\section{Methodology}

\begin{figure}[t]
    \centering
    \includegraphics[width = 0.8\textwidth]{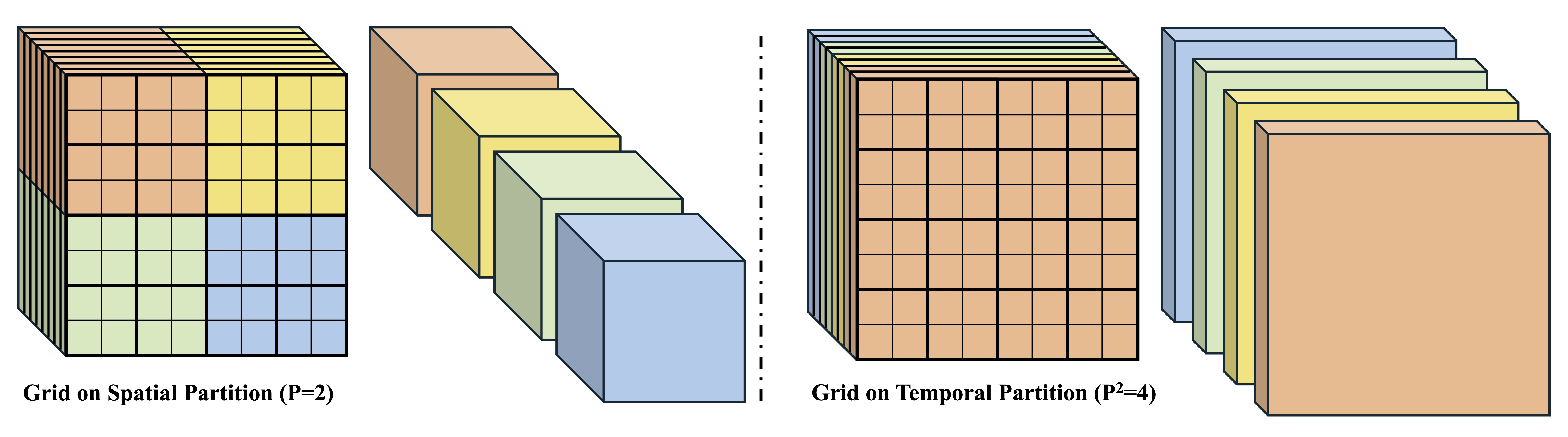}
    \caption{3D attention with grid. Our model significantly reduce the computational complexity of 3D attention while effectively preserving the interaction of spatiotemporal information by dividing spatiotemporal blocks in a grid manner.}
    \label{fig:figure3}
\end{figure}

\subsection{Overall Architecture}
As illustrated in Figure~\ref{fig:figure2}, our proposed CascadeV model consists of two main cascaded components: a base T2V model and a latent diffusion-based VAE (LDM-VAE) decoder. The base T2V model generates latnet representations aligned with textual semantic information, which serves as a conditional input for LDM-VAE. The VAE decoder, comprising a latent diffusion model and a standard VAE, augments the base model's output with high-frequency details, achieving a $32\times$ decoding ratio.

More precisely, we first use the base T2V model to sample from noise $\varepsilon_{BM} \sim \mathcal{N}(0, \mathbf{I})$ and get $x_{BM} = \mathcal{F}_{BM}(\varepsilon_{BM}, c_{text}, t)$, where $\mathcal{F}_{BM}(\cdot)$ is the sampling process of the base T2V model, $c_{text}$ is the text condition and $t$ is time. Subsequently, $x_{BM}$ is upsampled and concatenated with $\varepsilon_{DiT} \sim \mathcal{N}(0, \mathbf{I})$ along the channel dimension to obtain $x_{in} = [\varepsilon_{DiT}, \hat{x}_{BM}]$, where $\hat{x}_{BM}$ is unsampled $x_{BM}$. Next, DiT performs the first stage of decoding to obtain $x_{DiT} = \mathcal{F}_{DiT}(x_{in}, t)$, which is then finally decoded by the VAE decoder to produce a high-quality video $x = \mathcal{D}(x_{DiT})$, where  $\mathcal{F}_{DiT}(\cdot)$ is the sampling process of DiT and $\mathcal{D}(\cdot)$ is the decoder of VAE.

\subsection{Base T2V Model}
As described in Section~\ref{sec:3.4}, the base T2V model can be an existing T2V or I2V model, such as Open-Spra-Plan v1.1.0 \cite{pku_yuan_lab_and_tuzhan_ai_etc_2024_10948109}, SVD \cite{blattmann2023stable}. Therefore, we place greater attention on LDM-VAE to verify the feasibility of the cascade solution.

\subsection{Latent Diffusion Model-based VAE}
To ensure the quality of the generated results, current mainstream latent diffusion models typically limit the compression ratio to 8:1 \cite{rombach2022high}. Therefore, we incorporate DiT into the VAE, leveraging the generative capabilities of diffusion models to compensate for information loss and enable higher compression ratio decoding of the latent representations produced by the base T2V model.
Specifically, we use the latents generated by the base T2V model as conditional inputs to DiT for the first stage of decoding. This stage employs diffusion models to recover features lost at high compression ratio, producing representations within the VAE latent space. Subsequently, these representations undergo the second decoding phase through a standard VAE to produce high-quality videos.
We adopted the DiT block architecture from \cite{chen2024pixart} and implemented the following modifications:

\paragraph{Grid-based 3D Attention.} Current methods that compute spatial and temporal attention independently often disrupt the continuity between temporal and spatial information, resulting in noticeable transitions between video frames and a lack of realism. As shown in Figure~\ref{fig:figure3}, we introduce a spatiotemporal alternating grid 3D attention mechanism to better integrate spatiotemporal information. 
This method significantly reduces the computational load of 3D attention by alternately dividing spatial and temporal information.

Let $X \in \mathbb{R}^{F \times H \times W \times D}$ denotes the input feature map. For methods that compute spatial and temporal attention separately, the computational complexity is $\mathcal{O}(FH^{2}W^{2}D) + \mathcal{O}(F^{2}HWD)$, with a computational difference of $HW/F$. Typically, $HW/F$ is significantly greater than 1, resulting in a substantial disparity in the amount of information interacting during each attention computation.
As illustrated in the Figure~\ref{fig:figure3}, when partitioning spatiotemporal blocks, we set the temporal dimension partition to be the square of the spatial dimension, ensuring that the complexity of each attention computation is uniformly $\mathcal{O}(\frac{F^{2}H^{2}W^{2}}{P^{2}}D)$. Compared to computing global attention directly, this approach reduces the computational complexity by $\frac{1}{p^{2}}$. By adjusting $P$, we can balance performance and high-resolution inputs. Furthermore, since each computation incorporates spatiotemporal information, this method better captures spatiotemporal correlations, thereby ensuring consistency between output frames.

\paragraph{Removal of Text Cross-attention.} As described in \cite{pernias2023wurstchen}, the base model generates latent representations rich in textual semantic information, while the subsequent VAE's role is to supplement the high-frequency details necessary for high-quality outputs. This phenomenon is more pronounced when the base model is trained in the VAE latent space, where it can directly output videos consistent with the prompt. Therefore, in this stage, we remove text cross-attention, enabling the model to focus more on high-frequency detail refinement while reducing computational complexity.

\begin{figure}[t]
    \centering
    \includegraphics[width = 0.8\textwidth]{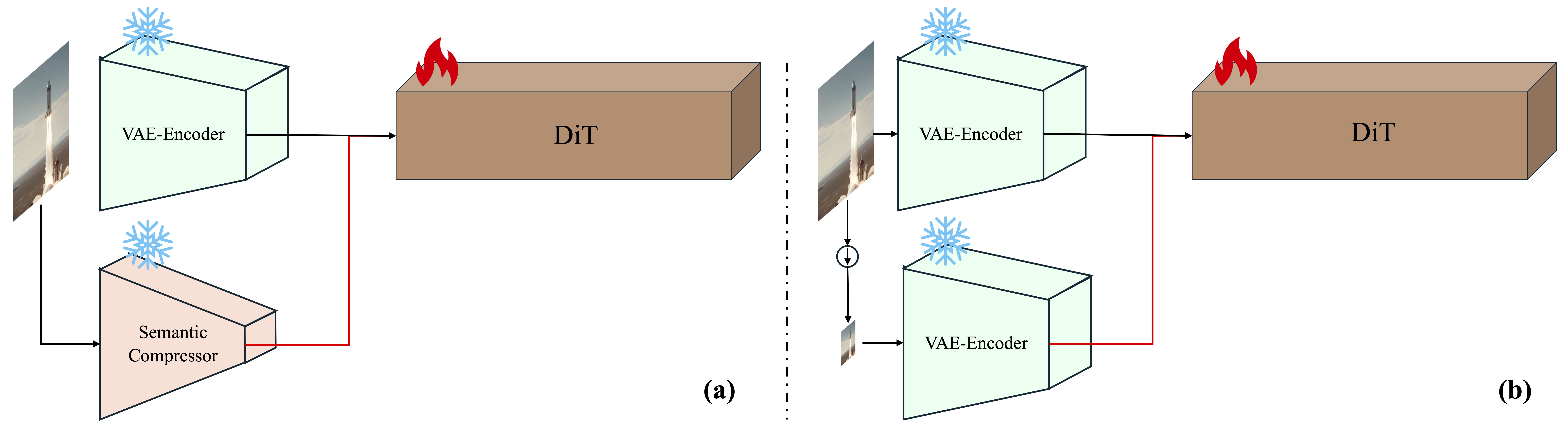}
    \caption{Two approaches for generating conditioning in DiT. (a) Utilizing high-compression Semantic compressor, which follows the \cite{pernias2023wurstchen} approach, results in a relative strong interdependence between the Base T2V Model and the DiT.
    (b) Implementing resize techniques to achieve higher compression ratio, which decouples the base model from the DiT, theoretically offering greater scalability.}
    \label{fig:figure4}
\end{figure}

\subsection{Connecting Base T2V Model and LDM-VAE}
\label{sec:3.4}
While the base model and LDM-VAE are trained independently, the inference stage requires LDM-VAE to process latent representations from the base model. These representations are then decoded to the standard VAE latent space for second-stage decoding, the transformation accomplished by DiT in the first stage.

Therefore, we propose two methods for aligning the latent spaces in DiT training (illustrated in Figure~\ref{fig:figure4}):

i) Leveraging established concepts from \cite{pernias2023wurstchen}, we compress the video at ratios of $8:1$ and $32:1$ using a standard VAE and Semantic Compressor \cite{pernias2023wurstchen} to obtain $x_{8:1}$ and $x_{32:1}$, respectively. Subsequently, $x_{32:1}$ serves as conditional input to DiT for decoding from the base model latent space to the VAE latent space.

ii) Alternatively, as shown in Figure~\ref{fig:figure4}(b), we first downsample the original video $x$ 4-fold via resizing to obtain $x_{4:1}$. Subsequently, a standard VAE compresses both $x$ and $x_{4:1}$, achieving $8:1$ and $32:1$ compression, respectively. In this configuration, the base model and LDM-VAE share the same latent space, with DiT functioning more like a latent space super-resolution model. This approach offers a significant advantage: since the latent spaces are naturally aligned, our DiT can be cascaded into existing T2V model decoders without fine-tuning, improving the quality of the results. Notably, it enables post-processing of pixel-space videos generated by T2V models. By encoding these videos with the VAE and inputting them into our DiT, we can enhance resolution and FPS without modifying the original model architecture.
\section{Experiments}

\subsection{Set Up}
One of the current challenges in high-quality video generation is the scarcity of high-quality open-source video datasets. We conducte experiments using the Intern4k dataset \cite{stergiou2022adapool}, which comprises 1000 high-quality videos with 4K resolution and 60 FPS and compare CascadeV's reconstruction capabilities with existing state-of-the-art open-source video models' VAEs, including Open-Sora-Plan v1.1 \cite{pku_yuan_lab_and_tuzhan_ai_etc_2024_10948109}, EasyAnimate v3 \cite{xu2024easyanimatehighperformancelongvideo}, and StableCascade \cite{pernias2023wurstchen}. 

Two distinct sets of metrics are used to evaluate the reconstruction performance of models: traditional image quality metrics, including Peak Signal-to-Noise Ratio (PSNR), Structural Similarity (SSIM), and Learned Perceptual Image Patch Similarity (LPIPS), as well as video quality metrics from VBench \cite{huang2023vbench}, which include Subject Consistency, Background Consistency, Temporal Flickering, Motion Smoothness, Imaging Quality, and Aesthetic Quality. 

\subsection{Quantitative Results}
As shown in Table \ref{tab:tra_metric}, our model, despite having a high compression ratio, generates numerous reasonable high-frequency details that may not precisely align with the ground truth in order to ensure effective reconstruction. Consequently, the reconstruction performance of our model on traditional metrics inferior to EasyAnimate v3 \cite{xu2024easyanimatehighperformancelongvideo} and Open-Sora-Plan v1.1.0 \cite{pku_yuan_lab_and_tuzhan_ai_etc_2024_10948109}, which employ lower compression ratios. Notably, our model and StableCascade \cite{pernias2023wurstchen} utilize a higher compression ration ($1024:1$) while still maintaining competitive performance, underscoring the efficacy of the cascade approach under high compression ratios.
\begin{table}[]
    \centering
    \setlength{\abovecaptionskip}{0.2cm}
    \begin{tabular}{lccc}
    \toprule
       Model / Compression Factor  & PSNR $\uparrow$& SSIM $\uparrow$ & LPIPS $\uparrow$ \\
    \midrule
    Open-Sora-Plan v1.1.0 / $(4 \times 8 \times 8 = 256)$ & \underline{25.7282} &  \underline{0.8000} &  \underline{0.1030} \\
    EasyAnimate v3 / $(4 \times 8 \times8 = 256)$ & \textbf{28.8666} & \textbf{0.8505} & \textbf{0.0818} \\
    StableCascade / $(1 \times 32 \times 32 = 1024)$ & 24.3336 & 0.6896 & 0.1395 \\
    Ours / $(1 \times 32 \times 32 = 1024)$ & 23.7320 & 0.6742 & 0.1786 \\
    \bottomrule
    \end{tabular}
    \caption{Quantitative results in traditional metrics.}
    \vspace{10pt}
    \label{tab:tra_metric}
\end{table}

\begin{table}[htbp]
    \centering
    \setlength{\abovecaptionskip}{0.2cm}
    \begin{tabular}{lcccccc}
    \toprule
       \makecell{Model \\ (Compression Factor)}  & \makecell{Subject \\ Consistency }& \makecell{Background \\ Consistency} & \makecell{Temporal \\ Flickering} & \makecell{Motion \\ Smoothness} & \makecell{Imaging \\ Quality} & \makecell{Aesthetic \\ Quality}\\
    \midrule
    \makecell{Open-Sora-Plan v1.1.0 \\ $(4 \times 8 \times 8 = 256)$ }& 0.9519 & 0.9618 & 0.9573 & 0.9789 & \underline{0.6791} & 0.5450 \\
    \makecell{EasyAnimate v3 \\ $(4 \times 8 \times8 = 256)$} & \underline{0.9578} & \textbf{0.9695} & \underline{0.9615} & \textbf{0.9845} & 0.6735 & 0.5535 \\
    \makecell{ StableCascade \\ $(1 \times 32 \times 32 = 1024)$ } & 0.9490 & 0.9517 & 0.9430 & 0.9639 & \textbf{0.6811} & \textbf{0.5675} \\
    \makecell{Ours \\ $(1 \times 32 \times 32 = 1024)$} & \textbf{0.9601} & \underline{0.9679} & \textbf{0.9626} & \underline{0.9837} & 0.6747 & \underline{0.5579} \\
    \bottomrule
    \end{tabular}
    \caption{Quantitative results in video quality metrics of VBench \cite{huang2023vbench}}.
    \vspace{10pt}
    \label{tab:vebnch}
\end{table}

To better assess the quality of reconstructed videos, we employed video quality evaluation metrics of VBench \cite{huang2023vbench}. Table \ref{tab:vebnch} demonstrates that our model performs well across various indicators, with particularly notable advantages in subject consistency, background consistency, and temporal flickering. These results validate the effectiveness of our proposed grid-based 3D attention. Considering our model's higher compression ratio, its ability to maintain leading or near-optimal performance across most metrics highlights the superiority of our approach.

\subsection{Qualitative Results}
\begin{figure}
    \centering
    \includegraphics[width=\linewidth]{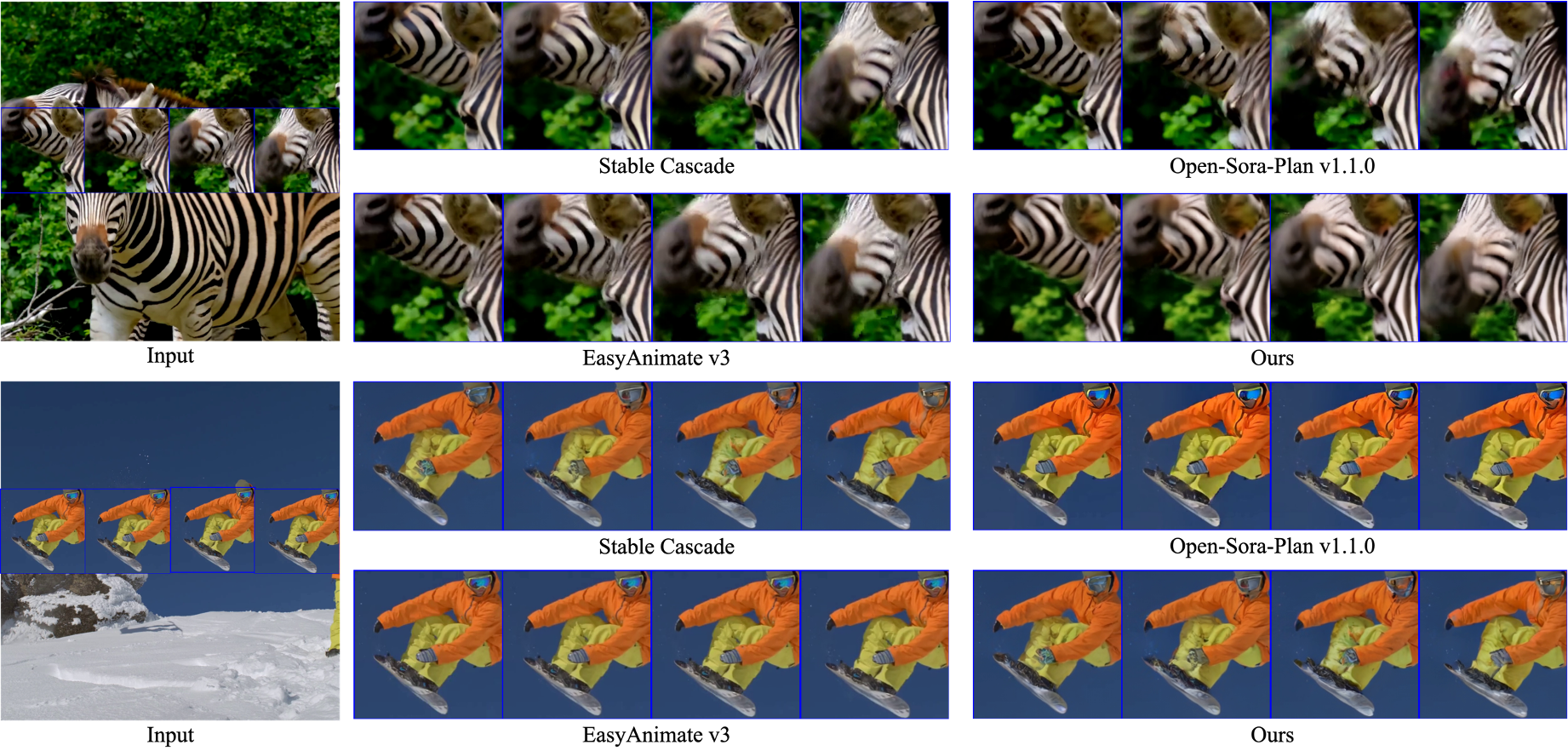}
    \caption{Qualitative results. Despite the higher compression ratio of our model, the reconstructed results still exhibit sufficient detail and good temporal consistency.}
    \label{fig:figure5}
\end{figure}

As illustrated in the Figure \ref{fig:figure5}, our model demonstrates a level of detail similar to, or even exceeding, that of VAEs with lower compression ratios (e.g., Open-Sora-Plan v1.1.0 \cite{pku_yuan_lab_and_tuzhan_ai_etc_2024_10948109}, EasyAnimate v3 \cite{xu2024easyanimatehighperformancelongvideo}). However, due to the higher compression ratio, more information is lost, which may result in discrepancies between the supplemented details in the reconstruction and the original input. Additionally, when comparing frames, the details supplemented by our model show better temporal consistency, ensuring the quality of the reconstructed video. This advantage is particularly pronounced when evaluated in video format.

\subsection{Cascading LDM-VAE with Existing Models}
Given that LDM-VAE inherently possesses certain generative capabilities, allowing it to supplement missing information during the reconstruction process, as articulated in Section \ref{sec:3.4}, by combining the VAEs used in the LDM training process, we can endow LDM-VAE with super-resolution or frame interpolation abilities. This enables the completion of tasks such as 4x video super-resolution or frame interpolation. 
As shown in Figure \ref{fig:figure1}, by employing the same VAE during the training process, we increase the resolution of the Open-Sora-Plan v1.1.0 generation results to 2K. Similarly, by employing VAEs with different temporal compression rates during training, we enhance the FPS of SVD-generated results to 28.
\section{Conclusion}
In this paper, we present CascadeV, a model based on the Würstchen architecture for high-quality T2V generation . Our model achieves a $32:1$ compression ratio in the latent space through a cascaded structure, significantly reducing the computational requirements for video generation. Moreover, we introduce a grid-based 3D attention mechanism that jointly processes spatial and temporal information, ensuring more natural transitions between generated video frames. Furthermore, our model can be cascaded with existing mature T2V models, enabling a 4$\times$ increase in resolution or FPS. In future work, we plan to explore further applications of our model in various video generation-related domains.

\section*{Acknowledgement}
We extend our gratitude to all contributors of the algorithms utilized in this paper, as their open-source code has been instrumental in facilitating our research. We also wish to express our appreciation to the Huggingface Community for providing convenient tools for model deployment and utilization.

\bibliographystyle{unsrt}
\bibliography{references}

\begin{thebibliography}{10}

\bibitem{pku_yuan_lab_and_tuzhan_ai_etc_2024_10948109}
PKU-Yuan Lab and Tuzhan~AI etc.
\newblock Open-sora-plan, April 2024.

\bibitem{blattmann2023stable}
Andreas Blattmann, Tim Dockhorn, Sumith Kulal, Daniel Mendelevitch, Maciej Kilian, Dominik Lorenz, Yam Levi, Zion English, Vikram Voleti, Adam Letts, et~al.
\newblock Stable video diffusion: Scaling latent video diffusion models to large datasets.
\newblock {\em arXiv preprint arXiv:2311.15127}, 2023.

\bibitem{ho2020denoising}
Jonathan Ho, Ajay Jain, and Pieter Abbeel.
\newblock Denoising diffusion probabilistic models.
\newblock {\em Advances in neural information processing systems}, 33:6840--6851, 2020.

\bibitem{song2020score}
Yang Song, Jascha Sohl-Dickstein, Diederik~P Kingma, Abhishek Kumar, Stefano Ermon, and Ben Poole.
\newblock Score-based generative modeling through stochastic differential equations.
\newblock {\em arXiv preprint arXiv:2011.13456}, 2020.

\bibitem{yang2023diffsound}
Dongchao Yang, Jianwei Yu, Helin Wang, Wen Wang, Chao Weng, Yuexian Zou, and Dong Yu.
\newblock Diffsound: Discrete diffusion model for text-to-sound generation.
\newblock {\em IEEE/ACM Transactions on Audio, Speech, and Language Processing}, 2023.

\bibitem{huang2023make}
Rongjie Huang, Jiawei Huang, Dongchao Yang, Yi~Ren, Luping Liu, Mingze Li, Zhenhui Ye, Jinglin Liu, Xiang Yin, and Zhou Zhao.
\newblock Make-an-audio: Text-to-audio generation with prompt-enhanced diffusion models.
\newblock In {\em International Conference on Machine Learning}, pages 13916--13932. PMLR, 2023.

\bibitem{hoogeboom2023simple}
Emiel Hoogeboom, Jonathan Heek, and Tim Salimans.
\newblock simple diffusion: End-to-end diffusion for high resolution images.
\newblock In {\em International Conference on Machine Learning}, pages 13213--13232. PMLR, 2023.

\bibitem{yang2024inf}
Zhuoyi Yang, Heyang Jiang, Wenyi Hong, Jiayan Teng, Wendi Zheng, Yuxiao Dong, Ming Ding, and Jie Tang.
\newblock Inf-dit: Upsampling any-resolution image with memory-efficient diffusion transformer.
\newblock {\em arXiv preprint arXiv:2405.04312}, 2024.

\bibitem{esser2024scaling}
Patrick Esser, Sumith Kulal, Andreas Blattmann, Rahim Entezari, Jonas M{\"u}ller, Harry Saini, Yam Levi, Dominik Lorenz, Axel Sauer, Frederic Boesel, et~al.
\newblock Scaling rectified flow transformers for high-resolution image synthesis.
\newblock In {\em Forty-first International Conference on Machine Learning}, 2024.

\bibitem{betker2023improving}
James Betker, Gabriel Goh, Li~Jing, Tim Brooks, Jianfeng Wang, Linjie Li, Long Ouyang, Juntang Zhuang, Joyce Lee, Yufei Guo, et~al.
\newblock Improving image generation with better captions.
\newblock {\em Computer Science. https://cdn. openai. com/papers/dall-e-3. pdf}, 2(3):8, 2023.

\bibitem{gupta2023photorealistic}
Agrim Gupta, Lijun Yu, Kihyuk Sohn, Xiuye Gu, Meera Hahn, Li~Fei-Fei, Irfan Essa, Lu~Jiang, and Jos{\'e} Lezama.
\newblock Photorealistic video generation with diffusion models.
\newblock {\em arXiv preprint arXiv:2312.06662}, 2023.

\bibitem{rombach2022high}
Robin Rombach, Andreas Blattmann, Dominik Lorenz, Patrick Esser, and Bj{\"o}rn Ommer.
\newblock High-resolution image synthesis with latent diffusion models.
\newblock In {\em Proceedings of the IEEE/CVF conference on computer vision and pattern recognition}, pages 10684--10695, 2022.

\bibitem{pernias2023wurstchen}
Pablo Pernias, Dominic Rampas, Mats~Leon Richter, Christopher Pal, and Marc Aubreville.
\newblock W{\"u}rstchen: An efficient architecture for large-scale text-to-image diffusion models.
\newblock In {\em The Twelfth International Conference on Learning Representations}, 2023.

\bibitem{peebles2023scalable}
William Peebles and Saining Xie.
\newblock Scalable diffusion models with transformers.
\newblock In {\em Proceedings of the IEEE/CVF International Conference on Computer Vision}, pages 4195--4205, 2023.

\bibitem{song2020denoising}
Jiaming Song, Chenlin Meng, and Stefano Ermon.
\newblock Denoising diffusion implicit models.
\newblock {\em arXiv preprint arXiv:2010.02502}, 2020.

\bibitem{zhang2022fast}
Qinsheng Zhang and Yongxin Chen.
\newblock Fast sampling of diffusion models with exponential integrator.
\newblock In {\em NeurIPS 2022 Workshop on Score-Based Methods}, 2022.

\bibitem{lu2022dpm}
Cheng Lu, Yuhao Zhou, Fan Bao, Jianfei Chen, Chongxuan Li, and Jun Zhu.
\newblock Dpm-solver: A fast ode solver for diffusion probabilistic model sampling in around 10 steps.
\newblock {\em Advances in Neural Information Processing Systems}, 35:5775--5787, 2022.

\bibitem{hu2021lora}
Edward~J Hu, Phillip Wallis, Zeyuan Allen-Zhu, Yuanzhi Li, Shean Wang, Lu~Wang, Weizhu Chen, et~al.
\newblock Lora: Low-rank adaptation of large language models.
\newblock In {\em International Conference on Learning Representations}, 2021.

\bibitem{zhang2023adding}
Lvmin Zhang, Anyi Rao, and Maneesh Agrawala.
\newblock Adding conditional control to text-to-image diffusion models.
\newblock In {\em Proceedings of the IEEE/CVF International Conference on Computer Vision}, pages 3836--3847, 2023.

\bibitem{ho2022cascaded}
Jonathan Ho, Chitwan Saharia, William Chan, David~J Fleet, Mohammad Norouzi, and Tim Salimans.
\newblock Cascaded diffusion models for high fidelity image generation.
\newblock {\em Journal of Machine Learning Research}, 23(47):1--33, 2022.

\bibitem{podell2023sdxl}
Dustin Podell, Zion English, Kyle Lacey, Andreas Blattmann, Tim Dockhorn, Jonas M{\"u}ller, Joe Penna, and Robin Rombach.
\newblock Sdxl: Improving latent diffusion models for high-resolution image synthesis.
\newblock {\em arXiv preprint arXiv:2307.01952}, 2023.

\bibitem{ronneberger2015u}
Olaf Ronneberger, Philipp Fischer, and Thomas Brox.
\newblock U-net: Convolutional networks for biomedical image segmentation.
\newblock In {\em Medical image computing and computer-assisted intervention--MICCAI 2015: 18th international conference, Munich, Germany, October 5-9, 2015, proceedings, part III 18}, pages 234--241. Springer, 2015.

\bibitem{yan2021videogpt}
Wilson Yan, Yunzhi Zhang, Pieter Abbeel, and Aravind Srinivas.
\newblock Videogpt: Video generation using vq-vae and transformers.
\newblock {\em arXiv preprint arXiv:2104.10157}, 2021.

\bibitem{wu2022nuwa}
Chenfei Wu, Jian Liang, Lei Ji, Fan Yang, Yuejian Fang, Daxin Jiang, and Nan Duan.
\newblock N{\"u}wa: Visual synthesis pre-training for neural visual world creation.
\newblock In {\em European conference on computer vision}, pages 720--736. Springer, 2022.

\bibitem{weissenborn2019scaling}
Dirk Weissenborn, Oscar T{\"a}ckstr{\"o}m, and Jakob Uszkoreit.
\newblock Scaling autoregressive video models.
\newblock In {\em International Conference on Learning Representations}, 2019.

\bibitem{yu2022generating}
Sihyun Yu, Jihoon Tack, Sangwoo Mo, Hyunsu Kim, Junho Kim, Jung-Woo Ha, and Jinwoo Shin.
\newblock Generating videos with dynamics-aware implicit generative adversarial networks.
\newblock In {\em 10th International Conference on Learning Representations, ICLR 2022}. International Conference on Learning Representations, ICLR, 2022.

\bibitem{wang2020g3an}
Yaohui Wang, Piotr Bilinski, Francois Bremond, and Antitza Dantcheva.
\newblock G3an: Disentangling appearance and motion for video generation.
\newblock In {\em Proceedings of the IEEE/CVF Conference on Computer Vision and Pattern Recognition}, pages 5264--5273, 2020.

\bibitem{tian2020good}
Yu~Tian, Jian Ren, Menglei Chai, Kyle Olszewski, Xi~Peng, Dimitris~N Metaxas, and Sergey Tulyakov.
\newblock A good image generator is what you need for high-resolution video synthesis.
\newblock In {\em International Conference on Learning Representations}, 2020.

\bibitem{ge2022long}
Songwei Ge, Thomas Hayes, Harry Yang, Xi~Yin, Guan Pang, David Jacobs, Jia-Bin Huang, and Devi Parikh.
\newblock Long video generation with time-agnostic vqgan and time-sensitive transformer.
\newblock In {\em European Conference on Computer Vision}, pages 102--118. Springer, 2022.

\bibitem{hong2022cogvideo}
Wenyi Hong, Ming Ding, Wendi Zheng, Xinghan Liu, and Jie Tang.
\newblock Cogvideo: Large-scale pretraining for text-to-video generation via transformers.
\newblock In {\em The Eleventh International Conference on Learning Representations}, 2022.

\bibitem{kondratyuk2023videopoet}
Dan Kondratyuk, Lijun Yu, Xiuye Gu, Jos{\'e} Lezama, Jonathan Huang, Rachel Hornung, Hartwig Adam, Hassan Akbari, Yair Alon, Vighnesh Birodkar, et~al.
\newblock Videopoet: A large language model for zero-shot video generation.
\newblock {\em arXiv preprint arXiv:2312.14125}, 2023.

\bibitem{ho2022imagen}
Jonathan Ho, William Chan, Chitwan Saharia, Jay Whang, Ruiqi Gao, Alexey Gritsenko, Diederik~P Kingma, Ben Poole, Mohammad Norouzi, David~J Fleet, et~al.
\newblock Imagen video: High definition video generation with diffusion models.
\newblock {\em arXiv preprint arXiv:2210.02303}, 2022.

\bibitem{singer2022make}
Uriel Singer, Adam Polyak, Thomas Hayes, Xi~Yin, Jie An, Songyang Zhang, Qiyuan Hu, Harry Yang, Oron Ashual, Oran Gafni, et~al.
\newblock Make-a-video: Text-to-video generation without text-video data.
\newblock {\em arXiv preprint arXiv:2209.14792}, 2022.

\bibitem{blattmann2023align}
Andreas Blattmann, Robin Rombach, Huan Ling, Tim Dockhorn, Seung~Wook Kim, Sanja Fidler, and Karsten Kreis.
\newblock Align your latents: High-resolution video synthesis with latent diffusion models.
\newblock In {\em Proceedings of the IEEE/CVF Conference on Computer Vision and Pattern Recognition}, pages 22563--22575, 2023.

\bibitem{ge2023preserve}
Songwei Ge, Seungjun Nah, Guilin Liu, Tyler Poon, Andrew Tao, Bryan Catanzaro, David Jacobs, Jia-Bin Huang, Ming-Yu Liu, and Yogesh Balaji.
\newblock Preserve your own correlation: A noise prior for video diffusion models.
\newblock In {\em Proceedings of the IEEE/CVF International Conference on Computer Vision}, pages 22930--22941, 2023.

\bibitem{he2022latent}
Yingqing He, Tianyu Yang, Yong Zhang, Ying Shan, and Qifeng Chen.
\newblock Latent video diffusion models for high-fidelity long video generation.
\newblock {\em arXiv preprint arXiv:2211.13221}, 2022.

\bibitem{yu2023video}
Sihyun Yu, Kihyuk Sohn, Subin Kim, and Jinwoo Shin.
\newblock Video probabilistic diffusion models in projected latent space.
\newblock In {\em Proceedings of the IEEE/CVF Conference on Computer Vision and Pattern Recognition}, pages 18456--18466, 2023.

\bibitem{ho2022video}
Jonathan Ho, Tim Salimans, Alexey Gritsenko, William Chan, Mohammad Norouzi, and David~J Fleet.
\newblock Video diffusion models.
\newblock {\em Advances in Neural Information Processing Systems}, 35:8633--8646, 2022.

\bibitem{zhang2023i2vgen}
Shiwei Zhang, Jiayu Wang, Yingya Zhang, Kang Zhao, Hangjie Yuan, Zhiwu Qin, Xiang Wang, Deli Zhao, and Jingren Zhou.
\newblock I2vgen-xl: High-quality image-to-video synthesis via cascaded diffusion models.
\newblock {\em arXiv preprint arXiv:2311.04145}, 2023.

\bibitem{chen2024pixart}
Junsong Chen, Chongjian Ge, Enze Xie, Yue Wu, Lewei Yao, Xiaozhe Ren, Zhongdao Wang, Ping Luo, Huchuan Lu, and Zhenguo Li.
\newblock Pixart-$\backslash$sigma: Weak-to-strong training of diffusion transformer for 4k text-to-image generation.
\newblock {\em arXiv preprint arXiv:2403.04692}, 2024.

\bibitem{stergiou2022adapool}
Alexandros Stergiou and Ronald Poppe.
\newblock Adapool: Exponential adaptive pooling for information-retaining downsampling.
\newblock {\em IEEE Transactions on Image Processing}, 32:251--266, 2022.

\bibitem{xu2024easyanimatehighperformancelongvideo}
Jiaqi Xu, Xinyi Zou, Kunzhe Huang, Yunkuo Chen, Bo~Liu, MengLi Cheng, Xing Shi, and Jun Huang.
\newblock Easyanimate: A high-performance long video generation method based on transformer architecture, 2024.

\bibitem{huang2023vbench}
Ziqi Huang, Yinan He, Jiashuo Yu, Fan Zhang, Chenyang Si, Yuming Jiang, Yuanhan Zhang, Tianxing Wu, Qingyang Jin, Nattapol Chanpaisit, Yaohui Wang, Xinyuan Chen, Limin Wang, Dahua Lin, Yu~Qiao, and Ziwei Liu.
\newblock {VBench}: Comprehensive benchmark suite for video generative models.
\newblock In {\em Proceedings of the IEEE/CVF Conference on Computer Vision and Pattern Recognition}, 2024.

\end{thebibliography}

\end{document}